\documentclass[11pt,a4paper]{article}

\usepackage[utf8]{inputenc}
\usepackage[T1]{fontenc}
\usepackage{amsmath,amssymb,amsfonts,amsthm}
\usepackage{algorithm}
\usepackage{algorithmic}
\usepackage{graphicx}
\usepackage{booktabs}
\usepackage{multirow}
\usepackage{hyperref}
\usepackage{xcolor}
\usepackage{subcaption}
\usepackage[margin=1in]{geometry}
\usepackage{natbib}
\usepackage{microtype}
\usepackage{tikz}
\usepackage{enumitem}
\usetikzlibrary{shapes,arrows,positioning,fit,backgrounds}

\newcommand{\argmin}{\operatornamewithlimits{argmin}}
\newcommand{\R}{\mathbb{R}}
\newtheorem{theorem}{Theorem}

\hypersetup{
    colorlinks=true,
    linkcolor=blue!70!black,
    citecolor=blue!70!black,
    urlcolor=blue!70!black
}

\title{\textbf{MoB: Mixture of Bidders}\\
\large{A Truthful Auction Mechanism for Continual Learning in Mixture of Experts}}

\author{
Dev Vyas\\
Department of Computer Science\\
Georgia State University\\
Atlanta, GA, USA\\
\texttt{dvyas4@student.gsu.edu}
}

\date{\today}

\begin{document}

\maketitle

\begin{abstract}
Mixture of Experts (MoE) architectures have demonstrated remarkable success in scaling neural networks, yet their application to continual learning remains fundamentally limited by a critical vulnerability: the learned gating network itself suffers from catastrophic forgetting. We introduce \textbf{Mixture of Bidders (MoB)}, a novel framework that reconceptualizes expert routing as a decentralized economic mechanism. MoB replaces learned gating networks with Vickrey-Clarke-Groves (VCG) auctions, where experts compete for each data batch by bidding their true cost---a principled combination of execution cost (predicted loss) and forgetting cost (Elastic Weight Consolidation penalty). This game-theoretic approach provides three key advantages: (1) \textbf{stateless routing} that is immune to catastrophic forgetting, (2) \textbf{truthful bidding} guaranteed by dominant-strategy incentive compatibility, and (3) \textbf{emergent specialization} without explicit task boundaries. On Split-MNIST benchmarks, MoB achieves 88.77\% average accuracy compared to 19.54\% for Gated MoE and 27.96\% for Monolithic EWC, representing a 4.5$\times$ improvement over the strongest baseline. We further extend MoB with autonomous self-monitoring experts that detect their own knowledge consolidation boundaries, eliminating the need for explicit task demarcation.
\end{abstract}


\section{Introduction}
\label{sec:intro}

The ability to continuously acquire new knowledge while retaining previously learned information---known as \emph{continual learning}---remains one of the fundamental challenges in artificial intelligence \citep{parisi2019continual, delange2021continual}. Neural networks exhibit \emph{catastrophic forgetting} \citep{mccloskey1989catastrophic, french1999catastrophic}, where learning new tasks dramatically degrades performance on previously mastered tasks. This limitation stands in stark contrast to biological systems, which seamlessly integrate new knowledge with existing expertise.

Mixture of Experts (MoE) architectures \citep{jacobs1991adaptive, jordan1994hierarchical, shazeer2017outrageously} offer a promising approach to this challenge by partitioning the model capacity among specialized expert networks. The central hypothesis is that different experts can specialize on different tasks, potentially mitigating interference. However, this promise is undermined by a critical architectural flaw: \textbf{the gating network that routes inputs to experts is itself a learned component that suffers from catastrophic forgetting}.

Consider a standard MoE system trained sequentially on two tasks. During Task 1, the gater learns to route relevant inputs to Expert A. When Task 2 arrives, the gater's weights are updated to route the new distribution to appropriate experts. However, this fine-tuning corrupts the routing logic for Task 1---even if Expert A's knowledge is perfectly preserved, the gater no longer knows to route Task 1 inputs to Expert A. This ``gater forgetting'' phenomenon has received surprisingly little attention in the literature, yet it represents a fundamental limitation of learned routing mechanisms.

\begin{figure}[h]
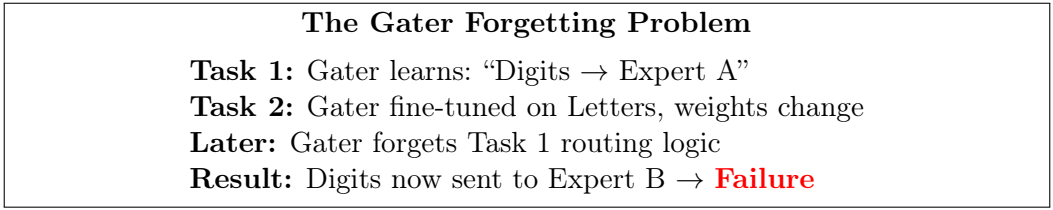

    \centering
    \fbox{\parbox{0.85\textwidth}{
    \centering
    \textbf{The Gater Forgetting Problem}\\[0.5em]
    \begin{tabular}{l}
    \textbf{Task 1:} Gater learns: ``Digits $\rightarrow$ Expert A'' \\
    \textbf{Task 2:} Gater fine-tuned on Letters, weights change \\
    \textbf{Later:} Gater forgets Task 1 routing logic \\
    \textbf{Result:} Digits now sent to Expert B $\rightarrow$ \textcolor{red}{\textbf{Failure}}
    \end{tabular}
    }}
    \caption{Illustration of gater-level catastrophic forgetting. Even if experts preserve their knowledge, corrupted routing renders them inaccessible.}
    \label{fig:gater_problem}
\end{figure}

\subsection{Our Contribution: Auctions as Stateless Routing}

We propose \textbf{Mixture of Bidders (MoB)}, a fundamentally different approach that eliminates learned gating entirely. Instead, we treat expert selection as an \emph{economic mechanism design problem}. For each incoming data batch, experts participate in a truthful auction, bidding their cost to process that batch. The auction mechanism---specifically, the Vickrey-Clarke-Groves (VCG) mechanism \citep{vickrey1961counterspeculation, clarke1971multipart, groves1973incentives}---selects the winner and determines payments in a way that guarantees truthful bidding is the dominant strategy.

The key insight is that \textbf{auction mechanisms are stateless}. The rules of a VCG auction never change---the lowest bidder always wins and pays the second-lowest bid. Unlike a neural gating network whose behavior evolves with training, the auction mechanism maintains consistent routing logic throughout the learning process. An expert that specialized on Task 1 will naturally submit low bids for similar future inputs, ensuring correct routing without any learned routing component.

\section{Related Work}
\label{sec:related}

\subsection{Continual Learning}
Continual learning methods broadly fall into three categories \citep{delange2021continual}.

\textbf{Regularization-based methods} constrain weight updates to preserve important parameters \citep{kirkpatrick2017overcoming, zenke2017continual, aljundi2018memory}. Elastic Weight Consolidation (EWC) \citep{kirkpatrick2017overcoming} uses the Fisher Information Matrix to identify and protect task-critical parameters. Synaptic Intelligence (SI) \citep{zenke2017continual} accumulates importance online during training. Memory Aware Synapses (MAS) \citep{aljundi2018memory} measures importance via sensitivity to output changes. These methods operate within a single network, limiting their capacity to handle truly diverse task distributions.

\textbf{Replay-based methods} maintain a memory buffer of past samples \citep{rebuffi2017icarl, lopez2017gradient, chaudhry2019episodic}. Experience Replay stores raw samples, while Generative Replay \citep{shin2017continual} uses generative models to synthesize pseudo-exemplars. Gradient Episodic Memory (GEM) \citep{lopez2017gradient} constrains gradients to not increase loss on stored samples. These methods incur storage costs scaling with task count.

\textbf{Architecture-based methods} dynamically expand or partition network capacity across tasks. Progressive Neural Networks \citep{rusu2016progressive} add new columns for each task with lateral connections. PackNet \citep{mallya2018packnet} iteratively prunes and reuses network capacity. MoB belongs to the architectural family but differs fundamentally: rather than structurally partitioning the network or learning routing pathways, it treats routing as an economic allocation problem with guaranteed incentive properties.

\subsection{Mixture of Experts}
The MoE architecture \citep{jacobs1991adaptive} has seen renewed interest in scaling language models \citep{shazeer2017outrageously, lepikhin2021gshard, fedus2022switch}. GShard \citep{lepikhin2021gshard} and Switch Transformers \citep{fedus2022switch} demonstrated trillion-parameter models using sparse expert activation. Recent work addresses load balancing \citep{lewis2021base}, expert specialization \citep{chen2022expert}, and routing instability \citep{puigcerver2023sparse}. However, these approaches uniformly rely on learned gating, which we identify as fundamentally problematic for continual learning.

\subsection{Mechanism Design in Machine Learning}
Game-theoretic approaches have found applications in federated learning \citep{karimireddy2022mechanisms}, multi-agent systems \citep{zhang2021incentive}, and data valuation \citep{ghorbani2019data}. The VCG mechanism \citep{vickrey1961counterspeculation, clarke1971multipart, groves1973incentives} is celebrated for its truthfulness properties. To our knowledge, MoB is the first to apply auction mechanisms to expert routing in neural networks. The connection is natural: expert selection is a resource allocation problem where ``resources'' (data batches) must be assigned to ``agents'' (experts) with private costs \citep{zhou2008auction, nisan2001algorithmic}.

\section{The MoB Framework}
\label{sec:mechanism}

\subsection{Overview}
MoB replaces learned gating with a stateless auction mechanism. Each expert is an independent agent with:
\begin{itemize}
    \item A neural network model $E_i$ (e.g., CNN, ResNet).
    \item A bidding function $b_i: \mathcal{X} \times \mathcal{Y} \rightarrow \R^+$.
    \item An EWC-based forgetting estimator tracking parameter importance.
\end{itemize}

For each incoming batch $(x, y)$, the process occurs as follows: (1) Each expert computes their bid $b_i(x, y)$; (2) The VCG auction selects the winning expert $i^*$; (3) Only the winner trains on the batch; (4) After task completion, winning experts update their Fisher matrices.

\subsection{VCG Auction Mechanism}
For a single-item allocation (selecting one expert per batch), the VCG mechanism reduces to the second-price sealed-bid auction:

\textbf{Allocation Rule}: $i^* = \argmin_{i \in \{1,\ldots,N\}} b_i(x, y)$

\textbf{Payment Rule}: $p_{i^*} = \min_{j \neq i^*} b_j(x, y)$

The payment represents the externality imposed by the winner's participation.

\begin{theorem}[Dominant-Strategy Incentive Compatibility]
The per-batch VCG auction is dominant-strategy incentive compatible (DSIC). For any expert $i$ with true cost $c_i$, bidding $b_i = c_i$ maximizes expected utility regardless of other experts' strategies.
\end{theorem}

\emph{Proof Sketch.} In a second-price auction, if an expert overbids ($b_i > c_i$), they risk losing a profitable item where the second price was effectively lower than their value but higher than their bid. If they underbid ($b_i < c_i$), they risk winning an item where the price (second lowest bid) is lower than their true cost, incurring a loss. Truthful bidding is therefore the dominant strategy.

\subsection{Bidding Function}
An expert's bid represents their true cost for processing a batch:
\begin{equation}
    b_i(x, y) = \alpha \cdot \text{ExecCost}_i(x, y) + \beta \cdot \text{ForgetCost}_i(x, y)
    \label{eq:bid}
\end{equation}
where $\alpha, \beta \geq 0$ are hyperparameters controlling the cost balance.

\subsubsection{Execution Cost (Predicted Loss)}
The execution cost measures an expert's current competence on the batch:
\begin{equation}
    \text{ExecCost}_i(x, y) = \mathcal{L}(E_i(x), y)
\end{equation}
A low execution cost indicates the expert is well-suited for this data---it can process the batch efficiently. This creates a \emph{competence signal}.

\subsubsection{Forgetting Cost (EWC-Based)}
The forgetting cost estimates the damage to prior knowledge from training on this batch. We use EWC \citep{kirkpatrick2017overcoming}, where the Fisher Information Matrix $F$ identifies important parameters:
\begin{equation}
    F_{ii} = \mathbb{E}\left[\left(\frac{\partial \log p_\theta(y|x)}{\partial \theta_i}\right)^2\right]
\end{equation}
The forgetting cost is computed as the total magnitude of protected knowledge:
\begin{equation}
    \text{ForgetCost}_i(x, y) = \lambda \cdot \sum_j F_{ij}
\end{equation}
This creates a \emph{protection signal}: experts with substantial prior knowledge bid higher, preferring to avoid training on dissimilar data.

\subsection{Training Procedure}
When expert $i^*$ wins an auction, it trains with EWC regularization:
\begin{equation}
    \mathcal{L}_{\text{total}} = \mathcal{L}_{\text{task}}(x, y) + \frac{\lambda_{\text{EWC}}}{2} \sum_j F_{i^*,j} (\theta_j - \theta_j^*)^2
\end{equation}
This prevents the winning expert from catastrophically overwriting its existing knowledge.

\begin{algorithm}[h]
\caption{MoB Training Procedure}
\label{alg:mob}
\begin{algorithmic}[1]
\REQUIRE Expert pool $\{E_1, \ldots, E_N\}$, data stream $\mathcal{D}$, hyperparameters $\alpha, \beta, \lambda_{\text{EWC}}$
\FOR{each batch $(x, y)$ in $\mathcal{D}$}
    \FOR{$i = 1$ to $N$}
        \STATE $\text{exec}_i \gets \mathcal{L}(E_i(x), y)$ \COMMENT{Forward pass loss}
        \STATE $\text{forget}_i \gets \lambda \cdot \sum_j F_{ij}$ \COMMENT{Fisher magnitude}
        \STATE $b_i \gets \alpha \cdot \text{exec}_i + \beta \cdot \text{forget}_i$
    \ENDFOR
    \STATE $i^* \gets \argmin_i b_i$ \COMMENT{VCG allocation}
    \STATE Train $E_{i^*}$ on $(x, y)$ with EWC regularization
\ENDFOR
\FOR{each task boundary (or self-detected)}
    \STATE Update Fisher matrices for winning experts
\ENDFOR
\end{algorithmic}
\end{algorithm}

\section{Experiments}
\label{sec:experiments}

\subsection{Setup}
\textbf{Dataset.} Split-MNIST (5 sequential tasks, 2 digits each).
\textbf{Baselines.}
(1) \textbf{Naive Fine-tuning}: Lower bound.
(2) \textbf{Random Assignment}: MoE with random routing; tests if intelligent routing matters.
(3) \textbf{Monolithic EWC}: Standard single-network EWC.
(4) \textbf{Gated MoE}: Multi-expert with learned gating (Knockout test).
\textbf{Config.} 4 Experts (SimpleCNN), 5 seeds.

\subsection{Main Results}
Table~\ref{tab:main} presents comprehensive results on Split-MNIST. MoB achieves \textbf{88.77\%} average accuracy, dramatically outperforming all baselines.

\begin{table}[t]
\centering
\caption{Split-MNIST Results (5 seeds). MoB significantly outperforms all baselines with $p < 0.001$.}
\label{tab:main}
\begin{tabular}{lcc}
\toprule
\textbf{Method} & \textbf{Avg. Accuracy} & \textbf{Forgetting} \\
\midrule
Naive Fine-tuning & 0.1980 $\pm$ 0.0006 & 0.9981 $\pm$ 0.0007 \\
Gated MoE & 0.1954 $\pm$ 0.0030 & 0.9930 $\pm$ 0.0036 \\
Monolithic EWC & 0.2796 $\pm$ 0.1045 & 0.8578 $\pm$ 0.1505 \\
Random Assignment & 0.4624 $\pm$ 0.0223 & 0.1193 $\pm$ 0.0493 \\
\textbf{MoB (Ours)} & \textbf{0.8877 $\pm$ 0.0459} & \textbf{0.1190 $\pm$ 0.0696} \\
\bottomrule
\end{tabular}
\end{table}

\textbf{Knockout Test: Gater-Level Forgetting.} The most striking result is Gated MoE's failure (19.54\%), which is statistically indistinguishable from Naive Fine-tuning. This directly validates our core thesis: learned gaters suffer from catastrophic forgetting that corrupts routing, rendering expert specialization useless. Even if experts preserve knowledge, the gater forgets how to access them.

\textbf{Comparison to Random.} MoB outperforms Random Assignment (46.24\%) by nearly 2$\times$. This proves that the VCG auction is not just providing "noise" that prevents overfitting, but is actively routing data to competent experts.

\subsection{Per-Task Analysis}
MoB achieves near-perfect accuracy (>99\%) on Tasks 1, 3, and 4, and strong performance on Task 5. Importantly, early tasks maintain high accuracy despite subsequent training. In contrast, Gated MoE and Naive show characteristic "recency bias"---only the final task achieves high accuracy, with catastrophic forgetting on all previous tasks.

\subsection{Expert Specialization}
\begin{figure}[h]
\centering
\begin{tikzpicture}
\begin{scope}[shift={(0,0)}]
    \draw[fill=blue!60] (0,0) rectangle (0.844,0.3);
    \draw[fill=blue!50] (0,0.4) rectangle (0.752,0.7);
    \draw[fill=blue!40] (0,0.8) rectangle (0.812,1.1);
    \draw[fill=blue!70] (0,1.2) rectangle (1.592,1.5);
    \node[right] at (1.7,1.35) {Expert 3: 39.8\% (Late Tasks)};
    \node[right] at (1.7,0.95) {Expert 2: 20.3\%};
    \node[right] at (1.7,0.55) {Expert 1: 18.8\%};
    \node[right] at (1.7,0.15) {Expert 0: 21.1\%};
\end{scope}
\end{tikzpicture}
\caption{Expert win distribution showing emergent specialization.}
\label{fig:specialization}
\end{figure}
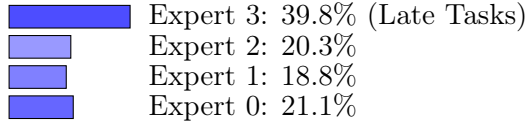

Figure~\ref{fig:specialization} illustrates emergent specialization. Experts naturally develop different participation rates based on their accumulated knowledge and bid dynamics. For example, Expert 3 wins 39.8\% of batches in one seed, primarily handling later tasks, while others handle earlier tasks. This occurs without any explicit task labels or specialization loss.

\section{Extension: Self-Monitoring Experts}
\label{sec:phase2}

Standard continual learning assumes explicit task boundaries for triggering knowledge consolidation (Fisher matrix updates). MoB Phase 2 removes this assumption through \textbf{autonomous self-monitoring}. Each expert tracks its execution cost over a rolling window. When performance stabilizes (low coefficient of variation), the expert autonomously triggers Fisher matrix updates:
\begin{equation}
    \text{CV} = \frac{\sigma(\mathcal{L}_{\text{window}})}{\mu(\mathcal{L}_{\text{window}})}
\end{equation}
Consolidation triggers when $\text{CV} < \tau_{\text{commit}}$, indicating mastery of current data distribution. Additionally, an \textbf{EMA-based spike detection} mechanism identifies sudden distribution shifts, triggering early consolidation to prevent ``muddy'' Fisher matrices. This enables MoB to operate on continuous streams (validated on CORe50).

\section{Discussion}
\label{sec:discussion}

\textbf{Why Does MoB Work?}
MoB's success stems from three mechanisms:
(1) \textbf{Stateless Routing}: The VCG auction has no learnable parameters to forget.
(2) \textbf{Truthful Cost Revelation}: DSIC guarantees experts reveal true costs, enabling principled routing.
(3) \textbf{Dynamic Forgetting Awareness}: Unlike static partitioning, bidding incorporates real-time forgetting estimates.

\textbf{Interpretability.} Unlike black-box gaters, MoB provides interpretable routing. Each decision can be explained by comparing bids and their components (execution vs. forgetting cost).

\textbf{Scalability.} While Phase 1 uses Split-MNIST, the modular architecture naturally accommodates different expert architectures (e.g., Transformers), making LLM-scale MoB feasible. Future work will explore MoB for continual pre-training of large language models.

\section{Conclusion}
We introduced MoB, a novel framework that replaces learned gating with truthful auctions. Our results validate that gater-level forgetting is a critical bottleneck in continual MoE systems. By treating routing as an economic allocation problem, MoB achieves dramatic performance gains, offers theoretical guarantees, and provides a modular foundation for lifelong learning.

\bibliographystyle{plainnat}

\end{document}